\documentclass{article}

\usepackage[numbers]{natbib} 
\usepackage[final]{neurips_2024}

\usepackage[utf8]{inputenc} 
\usepackage[T1]{fontenc}    
\usepackage{hyperref}       
\usepackage{url}            
\usepackage{booktabs}       
\usepackage{amsfonts}       
\usepackage{nicefrac}       
\usepackage{microtype}      
\usepackage{xcolor}         

\usepackage{pgffor}         
\usepackage{amsmath}        
\usepackage{amssymb}        
\usepackage{appendix}       
\usepackage{arydshln}       
\usepackage{mathtools}      
\usepackage{algorithm}      
\usepackage{algpseudocode}  
\usepackage{multirow}       
\usepackage[utf8]{inputenc} 
\usepackage[T1]{fontenc}    
\usepackage{graphicx}       
\usepackage{hyperref}       
\usepackage{url}            
\usepackage{booktabs}       
\usepackage{nicefrac}       
\usepackage{microtype}      
\usepackage{xcolor}         
\usepackage{kotex}          
\usepackage[figure,table]{hypcap}  

\newcommand{\systemName}{\text{ML$^2$Tuner}}

\newcommand{\paperTitle}{\systemName: Efficient Code Tuning via Multi-Level Machine Learning Models}

\title{\paperTitle}

\author{%
\hspace*{-0.6cm}
JooHyoung Cha$^{1}$ \quad  Munyoung Lee$^{2}$ \quad Jinse Kwon$^{2}$ \quad Jubin Lee$^{3}$ \quad Jemin Lee$^{2}$ \quad \textbf{Yongin Kwon}$^2$\thanks{Corresponding author.}\\
$^1$University of Science and Technology \quad \\ $^2$Electronics and Telecommunications Research Institute \quad $^3$Neubility\\
\texttt{\{jh.cha, munyounglee, kwonse, leejaymin, yongin.kwon\}@etri.re.kr}$^{1,2}$\\
\texttt{\{jubin0927\}@gmail.com}$^{3}$
}

\begin{document}

\maketitle

\vspace{-0.5cm}
\begin{abstract}

\vspace{-0.2cm}

The increasing complexity of deep learning models necessitates specialized hardware and software optimizations, particularly for deep learning accelerators. Existing autotuning methods often suffer from prolonged tuning times due to profiling invalid configurations, which can cause runtime errors. 
We introduce {\systemName}, a multi-level machine learning tuning technique that enhances autotuning efficiency by incorporating a validity prediction model to filter out invalid configurations and an advanced performance prediction model utilizing hidden features from the compilation process. Experimental results on an extended VTA accelerator demonstrate that {\systemName} achieves equivalent performance improvements using only 12.3\% of the samples required with a similar approach as TVM and reduces invalid profiling attempts by an average of 60.8\%, Highlighting its potential to enhance autotuning performance by filtering out invalid configurations

\end{abstract}


\section{Introduction}
\vspace{-0.2cm}

The increasing complexity of deep learning (DL) models has resulted in a significant increase in computational and memory demands~\cite{ai_memory_wall,epoch2023aitrends}, prompting growing attention to research on specialized hardware and software optimizations for efficient operation processing within resource constraints~\cite{ai_ha}.
In addition to widely recognized accelerators like Google's TPU~\cite{tpu_v4_google}, Nvidia's Tensor Core\cite{cuda_CUDA_Toolkit_nvidia}, and Intel's Gaudi~\cite{intel_gaudi3_2024}, a variety of DL accelerators are under active development. These accelerators are designed to specialize in DL computations, such as matrix-matrix multiplication, but to fully leverage the hardware's parallelism and maximize the use of internal memory, substantial software optimizations are required.


DL accelerator interfaces are typically divided into two types: high-level operator libraries~\cite{tensorflow_Abadi_TensorFlow_Large-scale_machine_2015,pytorch_Ansel_PyTorch_2_Faster_2024,onnxruntime_ONNX_Runtime_developers_ONNX_Runtime_2018,openvino,mxnet_chen2015mxnetflexibleefficientmachine} and low-level libraries~\cite{jax2018github,numba_llvm,eigenweb}. High-level operator libraries execute combinations of low-level code based on predefined rules that depend on operation inputs. While this approach is easy to apply, it has inherent optimization limitations, as it cannot define rules for all possible DL operations. In contrast, directly utilizing low-level libraries or generating machine codes can be challenging for users without in-depth hardware knowledge. DL compilers~\cite{xla_compiler,mlir,nest_compiler,halide_optimizing_parallelism} bridge this gap by taking DL operation inputs and autonomously generating optimized code.


Recent research on DL compiler has embraced machine learning-based autotuning techniques bypassing the requirement of domain expertise for reducing development time and the need for extensive human resources~\cite{AdaTune,rl_based_optimization_arco,dynatune_zhang2021}. Autotuning compensates for the lack of domain knowledge by exploring numerous code configurations and utilizing performance metrics to train a machine learning model for code optimization. 

Template-based methods, such as AutoTVM\cite{tvm_Chen2018TVMAA}, limit the search space, enhancing the chances of generating valid configurations but constraining the discovery of optimal solutions beyond the predefined space. 
On the other hand, template-free methods such as Ansor~\cite{zheng2020ansor} investigate a more extensive search space, which expands the possibility of finding better solutions but also increases the risk of generating invalid configurations by considering a wider range of options.

%


Irrespective of whether a template-based or template-free approach is used, identifying invalid configurations remains a significant challenge. The use of invalid configurations during machine learning model training can lead to reduced efficiency in the autotuning process and degrade the performance of the final optimized code~\cite{alt_failure,one_shot_tuner}.
The use of sophisticated back-end compilers, such as LLVM~\cite{llvm_papers}, helps minimize the probability of erroneous or invalid code generation on CPUs and GPUs. However, certain configurations can still lead to the generation of invalid code or sub-optimal performance in LLVM.
In contrast, DL accelerators that rely on scratchpad memory and lack the capacity for sophisticated back-end compilers often exhibit erroneous values or encounter runtime errors~\cite{heron_vta_contents_invalid,test_case_for_dlc,tvm_vta_failure_DiagNNose}.

To address the issue of prolonged tuning times caused by profiling invalid configurations, we propose {\systemName}, a multi-level machine learning tuning technique. 
While existing DL compilers attempt to reduce the number of invalid configurations, the diversity of hardware and architectures still leads to tuning delays and reduced cost model performance~\cite{HW_Aware_Initialization,NNSmith_for_TVM_TensorRT}.
By incorporating an additional validity prediction model focused solely on predicting configuration validity and training the performance prediction model exclusively with valid configurations, we can achieve more accurate performance predictions.
%
Furthermore, by integrating an advanced performance prediction model that considers hidden features generated during the compilation process.
Compared to TVM approach, {\systemName} achieves equivalent performance improvements across all ResNet18 layers while using only 12.3\% of the sample size and reducing the average number of invalid profiling attempts by 60.8\%. 

\section{System Design of \systemName}
\vspace{-0.1cm}
\begin{figure}[t]
    \centering
    \includegraphics[width=0.8\columnwidth]{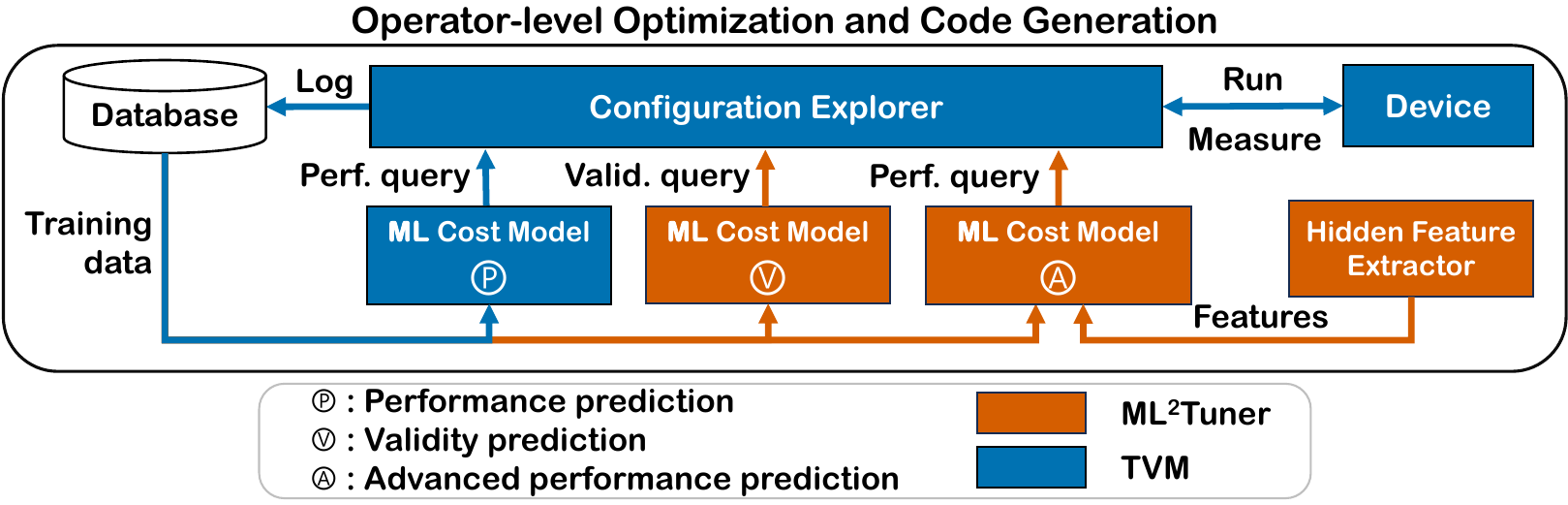}
    
    \caption{Overview of the automated tuning process of {\systemName}. Components in \textcolor{blue}{blue} represent the TVM's tuning process, while {\systemName} extends the components highlighted in
    \textcolor{orange}{orange}. }
    \label{fig:flow_of_nestuner}
    \vspace{-0.5cm}
\end{figure}
 
We introduce {\systemName}, an execution configuration tuner that employs multi-level machine learning, specifically designed for DL accelerators. 
Fig. ~\ref{fig:flow_of_nestuner} illustrates the overall automated tuning process of {\systemName}. 
While preserving the core components of TVM’s configuration explorer, ML cost model (P), and database, {\systemName} enhances functionality by incorporating two additional ML cost models (V and A), along with a hidden feature extractor to more effectively identify valid configurations and accurately predict performance.

\textbf{Configuration Explorer: } The configuration explorer selects $N$ code configurations with the highest potential for optimal performance to profile. It begins by defining the search space of all configurations for the given DL operation and assesses their potential based on predictions from the ML models. A code configuration can be a machine code or combination of low-level library calls.


\textbf{ML Model P: } Model P predicts the performance of configurations based on features provided by the configuration explorer, functioning similarly to the model employed in TVM. However, unlike TVM, which directly uses the configurations suggested by this model in subsequent profiling iterations, {\systemName} further evaluates these configurations using the models V and A.


\textbf{ML Model V: } Model V is dedicated to predicting the validity of configurations, utilizing the same features as Model P. Even if Model P predicts a configuration as highly optimal, {\systemName} avoids profiling it if Model V predicts it to be invalid. The configuration explorer iteratively applies models P and V until it accumulates $(\alpha + 1)\times N$ configurations. Once enough configurations are collected, {\systemName} moves on to the next step.

\textbf{Hidden Feature Extractor \& ML Model A: } From the $(\alpha + 1)\times N$ configurations selected—ranked by the highest predicted performance from Model P—Model A selects the final $N$ configurations expected to yield better performance.

Model A leverages not only the visible features used in Model P's predictions but also internal hidden features generated during the compilation process. During compilation, static analysis and optimization passes produce features such as branch decisions and loop size determinations. By incorporating these hidden features alongside the visible ones, {\systemName} builds a more precise performance prediction model without requiring extensive domain knowledge for hardware. Consequently, {\systemName} compiles all $(\alpha + 1)\times N$ configurations recommended by models P and V, extracts hidden features during this process, and uses model A to re-evaluate and select the final $N$ configurations.


\textbf{Profiling \& Training:} The final configurations selected by the configuration explorer are executed on real hardware. Validity is assessed by checking for crashes and verifying output correctness. For valid configurations, execution times are measured and logged into the database. In each iteration, $N$ configurations are executed; the validity results are used to train Model V, while the execution times of the valid configurations are used to train Models P and A.

\section{Experimental Results}
\vspace{-0.1cm}

\vspace{-0.2cm}
\textbf{Experimental Setup: } To validate the effectiveness of {\systemName}, we extended the open-source DL accelerator VTA~\cite{vta_moreau2019hardwaresoftwareblueprintflexibledeep} on a Xilinx ZCU102 board~\cite{ZCU102_evaluation}.
%
We also implemented the core components of {\systemName}, including the configuration explorer, hidden feature extractor, and the three ML cost models(P, V, and A), in the PyTorch Glow compiler~\cite{glow_kingma2018glowgenerativeflowinvertible} to compare the performance of
{\systemName} with that of TVM, which uses only the ML cost model P.
Additionally, we developed low-level libraries~\cite{ones_ai_VTA_LIB} for our VTA and implemented a back-end compiler~\cite{nest_compiler} to execute DL operations by integrating these components.
For machine learning-based autotuning, {\systemName} adopts XGBoost (v2.1.1)\cite{xgboost}. 
To identify and extract hidden features, {\systemName} uses an internally integrated compiler. It collects data such as iteration counts from configurations, values affected by conditional expressions, and variations resulting from branch statements. Furthermore, it captures details about the optimization and internal tiling strategies during the code generation process.

Our experiments targeted 10 types of convolution layers from ResNet18~\cite{resnet18}, trained on the ImageNet dataset, with the hyperparameters set to $N=10$ and $\alpha=1.0$.
The details of the experimental setup, such as the hyperparameters, are specified in Appendix~\ref{experimental_information}.

\begin{figure}[htbp]
    \centering
    \includegraphics[width=1.0\columnwidth]{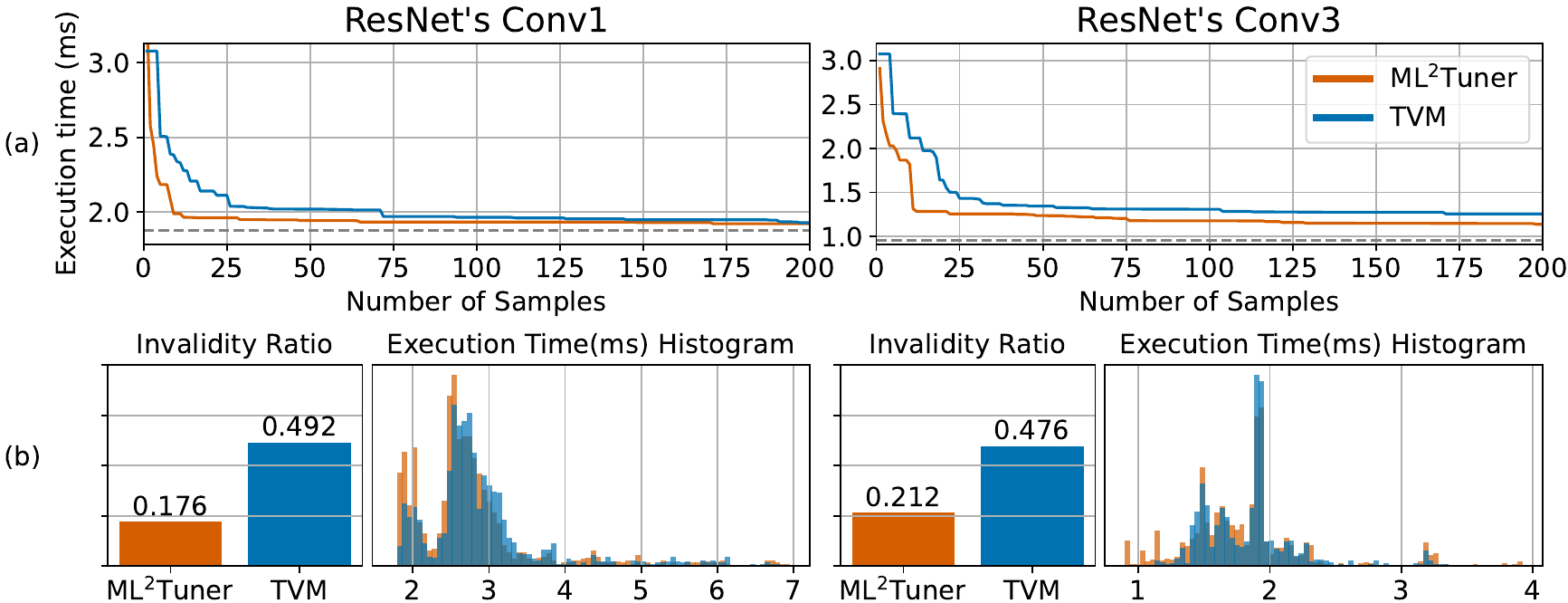}

    \vspace{-0.0cm}
    \caption{Experimental results for Conv1 and Conv2 of ResNet18. The \textcolor{orange}{orange} represents the results of {\systemName}, while the \textcolor{blue}{blue} represents the results of the TVM approach. (a) The x-axis shows the number of configurations tested during the tuning process, and the y-axis shows the lowest execution time among the cumulative configurations. (b) The left plot displays the invalidity ratio, while the right plot presents a normalized histogram of execution times for the valid configurations.}

    \label{fig:result_of_auto_tuning}
\end{figure}

\begin{figure}[bthp]
    \centering
    \includegraphics[width=0.7\columnwidth]{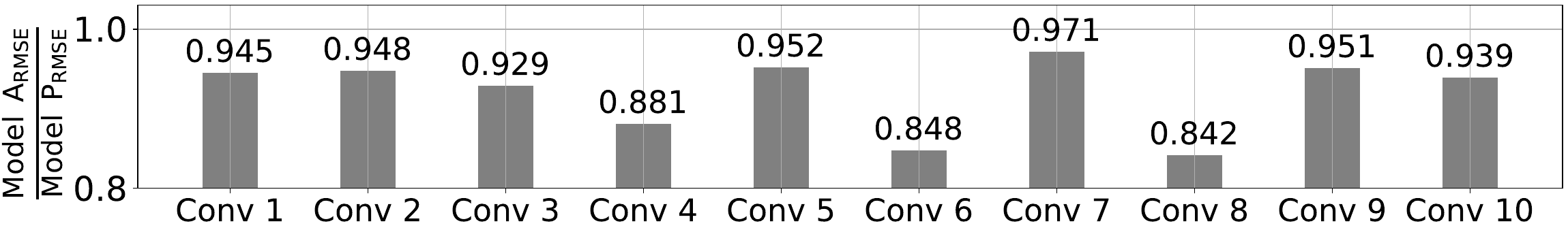}
    \vspace{-0.0cm}
    \caption{Ratio of RMSE values of model A compared to model P across different layers.} 
    \label{fig:compare_rmse}
    \vspace{-0.5cm}
\end{figure}


\textbf{Overall Performance: } Fig. ~\ref{fig:result_of_auto_tuning} (a) compares the tuning process of each ResNet18 layer using {\systemName} with the existing TVM method that employs a single machine learning model for performance prediction. Due to the random initialization of the machine learning models, results may vary with each attempt; therefore, we conducted 10 experiments and plotted the average values.

In our analysis, we identified the point of convergence in TVM's configuration exploration as the stage where the same value repeated more than 10 times. We then compared the average number of configuration explorations required by {\systemName} to reach equivalent performance. 
The results indicate that, for ResNet's Conv1 layer, {\systemName} achieved equivalent performance using only 11.2\% of the configurations attempted by TVM. For the Conv3 layer, this percentage was 11.3\%, with an average of 12.3\% across all ResNet18 layers.

\textbf{Distribution of configurations: } Fig. ~\ref{fig:result_of_auto_tuning} (b) shows the invalidity ratio of the configurations proposed by the machine learning models of {\systemName} and TVM during the tuning process and a normalized performance histogram for valid configurations. In preliminary experiments, random sampling for Conv1 yielded an invalidity ratio of 0.926. Using TVM's model reduced the invalidity ratio to 0.492, indicating a higher rate of selecting valid configurations than random chance. However, {\systemName} further reduced the invalidity ratio to 0.176, demonstrating greater efficiency in tuning time by avoiding invalid configurations. Similar trends were observed for Conv2 through Conv10. The histogram confirms that {\systemName} not only identifies valid configurations more effectively but also selects configurations skewed toward better performance, as indicated by the leftward shift.


\textbf{Impact of Hidden Features: } To demonstrate that ML model A predicts performance more accurately than ML model P by utilizing additional hidden features, we measured the Root Mean Square Error (RMSE) of both models. Fig. ~\ref{fig:compare_rmse} presents the RMSE ratio of model A compared to model P for each ResNet18 layer. The experimental results show an average ratio of 0.919, indicating that model A achieves lower prediction errors. Although extracting hidden features requires compiling $\alpha$ times more configurations, this investment yields more accurate performance predictions, which in turn improves the selection of final configurations. Appendix~\ref{hyper_f_score_normalized} lists the hidden features, highlighting those with high importance in performance prediction.

\section{Conclusion and Future Work}
\vspace{-0.1cm}

In this paper, we introduce {\systemName}, a multi-level machine learning tuning technique designed to enhance the efficiency and effectiveness of autotuning for DL accelerators. By incorporating a validity prediction model (Model V) to filter out invalid configurations and an advanced performance prediction model (Model A) that takes advantage of hidden features extracted during the compilation process, {\systemName} addresses key limitations of existing autotuning methods that rely on a single machine learning model.


As future work, we plan to evaluate {\systemName} on a diverse range of hardware platforms beyond VTA-style DL accelerators to assess its generalizability and effectiveness across various architectures. 
Additionally, we aim to incorporate advanced machine learning techniques, such as reinforcement learning and Bayesian optimization, to further enhance the tuning process. 
Furthermore, we intend to develop a self-recovering system capable of automatically handling runtime errors during tuning, thereby improving robustness and reducing the need for manual intervention. These efforts aim to enhance the adaptability, efficiency, and practicality of {\systemName} in optimizing DL models across various hardware environments.

\vspace{-0.3cm}
\section*{Acknowledgment}
\vspace{-0.3cm}
This work was supported by the Institute of Information \& Communications Technology Planning \& Evaluation(IITP) grant funded by the Korea government(MSIT) (No.RS-2024-00459797, Development of ML compiler framework for on-device AI), (No.RS-2023-00277060, Development of open edge AI SoC hardware and software platform) and (No.2022-0-00454, Technology development of smart edge device SW development platform).


\medskip

{\small
\bibliographystyle{unsrt}
\bibliography{egbib}
}

\newpage

\appendix

\section{Additional Experimental Information}
\label{experimental_information}

\subsection{Hardware} \label{appendix_hardware}



The extended VTA was implemented on the Xilinx ZCU FPGA, with the configuration details outlined in Table ~\ref{tab:evta_config}. FPGA boards, including the Intel DE10 Nano, Xilinx PNYQ, Xilinx ZCU104~\cite{zcu104_tvm_vta_configuration}, and Ultra96, are equipped with default VTA presets provided by TVM. However, no preset information was available for the Xilinx ZCU102 used in the experiments.

To proceed with the research on the Xilinx ZCU102, some adjustments were made to the VTA configuration based on ~\cite{tvm_vta_configuration}. First, the metadata of the VTA configurations, namely "Target" and "HW VER" were modified.
The following attributes were left unchanged: (1) LOG INP WIDTH, (2) LOG WGT WIDTH, (3) LOG ACC WIDTH, (4) LOG BATCH, and (5) LOG BLOCK.
The remaining four attributes were adjusted as follows: (1) LOG UOP BUFF SIZE, (2) LOG INP BUFF SIZE, (3) LOG WGT BUFF SIZE, and (4) LOG ACC BUFF SIZE, 
with each value increased by 1.

\begin{table}[h]
\caption{Parameters of VTA Configurations}
\resizebox{\columnwidth}{!}{%
\begin{tabular}{lclc}
\toprule
\multicolumn{1}{c}{Attribute} & Format & \multicolumn{1}{c}{Description} & Value \\
\midrule
TARGET & String & The TVM device target. & zcu102 \\
HW VER & String & VTA hardware version number. & 0.0.1 \\
LOG INP WIDTH & Int (log2) & Input data type signed integer width. & 3 \\
LOG WGT WIDTH & Int (log2) & Weight data type signed integer width. & 3 \\
LOG ACC WIDTH & Int (log2) & Accumulator data type signed integer width. & 5 \\
LOG BATCH & Int (log2) & VTA matrix multiply intrinsic input/output dimension 0. & 0 \\
LOG BLOCK & Int (log2) & VTA matrix multiply inner dimensions. & 4 \\
LOG UOP BUFF SIZE & Int (log2) & Micro-op on-chip buffer in Bytes. & 16 \\
LOG INP BUFF SIZE & Int (log2) & Input on-chip buffer in Bytes. & 16 \\
LOG WGT BUFF SIZE & Int (log2) & Weight on-chip buffer in Bytes. & 19 \\
LOG ACC BUFF SIZE & Int (log2) & Accumulator on-chip buffer in Bytes. & 18 \\
\bottomrule
\end{tabular}%
}
\label{tab:evta_config}
\end{table}

\subsection{Neural Network}



Table~\ref{tab:resnet18_combined} provides an analysis of the convolution layers of ResNet18 and highlights the ratio of invalid configurations in VTA. Part (a) provides detailed information about the parameters of each convolution layer, including the input shape, kernel filter, output shape, padding, and stride for each layer. Part (b) presents the invalidity ratio of configurations for each convolution layer.

\begin{table}[!h]
\centering
\caption{Profiling Layer of Convolution in ResNet18 and Ratio of Validity Configurations on VTA}
\begin{tabular}{lcccccc}
\toprule
\multirow{2}{*}{Name} & \multicolumn{4}{c}{(a) Information} & \multirow{2}{*}{(b) Invalidity Ratio} \\
\cmidrule(lr){2-5} 
& \multicolumn{1}{c}{H, W, C} & \multicolumn{1}{c}{KC, KH, KW} & \multicolumn{1}{c}{OH, OW} & \multicolumn{1}{c}{Pad, Stride} \\
\midrule
Conv 1 & 56,56,64 & 64,3,3 & 56,56 & 1,1 &  0.8264 \\
Conv 2 & 56,56,64 & 128,1,1   & 28,28 & 0,2  &  0.7966 \\ 
Conv 3 & 56,56,64 & 128,3,3   & 28,28 & 1,2  &  0.8057 \\ 
Conv 4 & 28,28,128 & 128,3,3  & 28,28 & 1,1  &  0.6935 \\ 
Conv 5 & 28,28,128 & 256,1,1  & 14,14 & 0,2  &  0.5249 \\
Conv 6 & 56,56,64 & 128,1,1   & 28,28 & 0,2  &  0.5249 \\
Conv 7 & 56,56,64 & 128,3,3   & 28,28 & 1,2  &  0.5249 \\
Conv 8 & 28,28,128 & 128,3,3  & 28,28 & 1,1  &  0.5047 \\
Conv 9 & 56,56,64 & 128,3,3   & 28,28 & 1,2  &  0.5047 \\
Conv 10 & 28,28,128 & 128,3,3 & 28,28 & 1,1  &  0.5047 \\
\bottomrule
\end{tabular}
\label{tab:resnet18_combined}
\end{table}

The optimizable features in our VTA implementation and backend compiler are based on tiling and the number of virtual threads. An invalid configuration is defined as one in which a calculation fails due to a register error, requiring a manual reboot, or a test fails because the result differs from the expected result. Conversely, a valid calculation is a configuration that successfully completes the task without errors.

\subsection{Experimental HyperParameter on \systemName}

Models P and A are designed to identify the highest predicted performance, while Model V is configured for classification tasks. Consequently, Models P, A, and V have distinct objective functions and loss functions, which lead to different search ranges for these parameters during hyperparameter tuning.
Models P and A were optimized using regression and ranking objectives to achieve the highest prediction performance. In contrast, Model V was optimized using binary classification objectives to define its search range.

To optimize the hyperparameters of XGBoost, we conducted a grid search~\cite{scikit-learn} based on the search space outlined in Table~\ref{tab:combined_hyperparameters}. 
The overall tuning results are presented in Table~\ref{tab:combined_validation_and_hyper_parameter}.

\begin{table}[h]
\centering
\caption{Exploration Range and Hyperparameters for XGBoost Models}
\hspace*{-0.5cm}
\resizebox{1.05\columnwidth}{!}{
\begin{tabular}{lcccc}
\toprule
\textbf{Parameter} & \textbf{Search Space} & \textbf{Model P} & \textbf{Model V} & \textbf{Model A} \\
\midrule
objective & -- & reg:squarederror & binary:hinge & reg:squarederror \\
boost round & -- & 300 & 300 & 300 \\
max depth & $ \{ x \in \mathbb{N} \mid 3 \leq x \leq 15 \} $ & 14 & 5 & 14 \\
min child weight & $ \{ x \in \mathbb{R} \mid 1 \leq x < 10 \} $ & 3 & 3 & 3 \\
gamma & $ \{ x \in \mathbb{R} \mid 0.0 \leq x < 1.0 \} $ & 0.0 & 0.0 & 0.0 \\
subsample & $ \{ x \in \mathbb{R} \mid 0.4 \leq x \leq 1.0 \} $ & 1.0 & 0.6 & 1.0 \\
colsample bytree & $ \{ x \in \mathbb{R} \mid 0.5 \leq x \leq 1.0 \} $ & 1.0 & 0.6 & 1.0 \\
learning rate & $ \{0.001, 0.01, 0.1, 0.2, 0.3\} $ & 0.01 & 0.1 & 0.01 \\
reg alpha & $ \{ x \in \{-5, -2, -1, 0, 1\} \mid 1 \times 10^{x} \} $ & $1 \times 10^{-5}$ & $1 \times 10^{-2}$ & $1 \times 10^{-5}$ \\
\bottomrule
\end{tabular}
}
\label{tab:combined_hyperparameters}
\end{table}

\section{Additional Experimental Results}

\subsection{Comparison of HyperParameter for Model V and Model A}


The accuracy and training time were calculated by changing the objective function. The time and accuracy for the 10 ResNet18 layers in VTA are shown in Table~\ref{tab:evta_config}.


Models P and A were configured using both Ranking and Regression objectives to compare their prediction performance, while Model V assessed the validity of configurations using both Binary classification and Regression objectives. For performance comparison, the hyperparameters are shown in Table~\ref{tab:combined_hyperparameters}.


As shown in Table~\ref{tab:combined_validation_and_hyper_parameter}, Model P and A achieved the highest prediction performance when using Regression objectives, outperforming Rank objectives by 0.06\%p in accuracy and being 1.70x faster in computation time. In Model V, the difference in computation time required for different objective functions was greater than the 
difference in accuracy.

\begin{table}[h]
\centering
\caption{Comparison of \systemName\,with objective function and loss}

\begin{tabular}{lclcc}
\toprule
Model & Objective Function & Loss & Accuracy &  Time (sec) \\
\midrule
\multirow{2}{*}{Model P and A} & Regression & Squared Error      & 99.55 & 320.21 \\
                                 & Rank~\cite{xgboost_rank}    & Logistic   & 99.49 & 537.74 \\
\midrule
\multirow{4}{*}{Model V} & \multirow{2}{*}{Regression} & Squared Error & 99.49 & 316.23 \\
                            &         & Logistic      & 99.47 & 350.89 \\
\cmidrule(lr){2-5}
                            & \multirow{2}{*}{Binary~\cite{binary_Imbalance-XGBoost}}  & Hinge         & 99.41 & 176.73 \\
                            &         &  Logistic      & 99.55 & 537.74 \\
\bottomrule
\end{tabular}
\label{tab:combined_validation_and_hyper_parameter}
\end{table}

\subsection{Impact of HyperParameters}

\label{hyper_f_score_normalized}

The layer and kernel information was not displayed during the evaluation of the visible features.
Table~\ref{tab:hyper_f_score_normalized} provides an overview of both visible and hidden features values. The visible features are highlighted in \textcolor{blue}{blue}, while the hidden features are presented in \textbf{black}. In addition, the impact of each hyperparameter is expressed as a percentage, offering insight into their relative importance and contribution to the model's performance.

Three parameters significantly influence the visible features and have a substantial impact on the model's execution time: Tile Width(TW), Tile Height(TH), and Number of Virtual Threads(nVT). The TW and TH parameters are used in tiling optimization to divide operations into smaller blocks. The value of nVT represents the number of virtual threads involved in parallel operations.


Hidden features consist of information whose values are derived from visible features or collected through internal branching mechanisms. These  features may include derived parameters or flags that influence execution paths. In cases where a feature name contains a Boolean operation, this means that the value of the feature is binary, either true or false, and represents the presence or absence of a particular condition or operation. If a Boolean operation is enclosed in parentheses within the feature name, it indicates that the variable's value depends on the specific branch taken during execution.

\begin{table}[!th]
\centering

\caption{Importance of Visible and Hidden Features}
\resizebox{\columnwidth}{!}{%
\begin{tabular}{lllllllllllll}
\toprule
\multicolumn{1}{c}{\multirow{2}{*}{Feature}} & \multicolumn{1}{c}{\multirow{2}{*}{GeoAVG}} & \multicolumn{10}{c}{Normalized Feature Importance Score (\%)} \\
\multicolumn{1}{c}{}                         & \multicolumn{1}{c}{}                        & \multicolumn{1}{c}{Conv1} & \multicolumn{1}{c}{Conv2} & \multicolumn{1}{c}{Conv3} & \multicolumn{1}{c}{Conv4} & \multicolumn{1}{c}{Conv5} & \multicolumn{1}{c}{Conv6} & \multicolumn{1}{c}{Conv7} & \multicolumn{1}{c}{Conv8} & \multicolumn{1}{c}{Conv9} & \multicolumn{1}{c}{Conv10} \\
\midrule
\textcolor{blue}{TW} & 29.268 & 19.685 & 27.412 & 23.810 & 25.779 & 32.631 & 30.633 & 28.432 & 31.596 & 31.975 & 32.582 \\
\textcolor{blue}{TH} & 25.925 & 15.256 & 19.768 & 20.186 & 22.288 & 27.235 & 29.335 & 29.201 & 32.406 & 31.028 & 30.622 \\
nVirtualThread > 0 (threadIdx) &  8.468 & 10.581 &  8.434 &  9.058 &  9.130 &  6.937 &  6.231 &  7.941 &  8.101 &  7.579 &  7.594 \\
\textcolor{blue}{nVT} &  8.194 &  5.413 &  7.907 &  7.505 &  7.519 &  7.965 &  7.269 &  7.941 &  8.912 &  9.237 &  8.574 \\
nFilterInLoop &  4.933 &  7.382 &  5.271 &  5.435 &  4.565 &  3.854 &  4.154 &  3.842 &  4.321 &  4.737 &  3.920 \\
sizeOutTileH &  4.083 &  2.215 &  3.163 &  3.364 &  3.491 &  4.625 &  4.673 &  4.611 &  4.861 &  4.737 &  4.655 \\
sizeOutTileW &  4.166 &  2.707 &  4.217 &  3.364 &  3.491 &  5.139 &  4.413 &  4.098 &  4.051 &  4.500 &  4.655 \\
nVirtualThread > 0 (threadIdx) 2 &  3.563 &  7.382 &  5.535 &  5.435 &  5.102 &  1.799 &  2.596 &  2.561 &  1.620 &  1.658 &  1.960 \\
sizeOutTileBoundaryW &  3.069 & 13.287 &  6.326 &  7.764 &  6.176 &  1.542 &  2.336 &  3.330 &  0.810 &  0.711 &  0.735 \\
outDummyH(b0!=0) &  1.946 &  0.738 &  1.581 &  1.812 &  2.685 &  2.569 &  2.596 &  2.561 &  1.080 &  1.184 &  1.715 \\
nFilterInLoop &  1.946 &  1.722 &  2.636 &  2.588 &  2.954 &  1.542 &  2.336 &  2.305 &  0.810 &  0.711 &  0.980 \\
resizedOutTileH(b0==0) &  1.233 &  9.104 &  3.426 &  3.623 &  1.611 &  0.257 &  0.260 &  0.512 &  0.270 &  0.237 &  0.245 \\
outDummyH(b0==0) &  1.069 &  1.476 &  1.845 &  3.623 &  2.148 &  0.771 &  0.779 &  1.025 &  0.270 &  0.237 &  0.245 \\
Kn / nFilterInLoop / nVirtualThread / 16 &  0.740 &  0.492 &  0.527 &  0.518 &  0.806 &  1.799 &  1.038 &  0.768 &  0.270 &  0.474 &  0.490 \\
sizeInTileW &  0.411 &  0.246 &  0.264 &  0.259 &  0.269 &  0.514 &  0.519 &  0.256 &  0.270 &  0.474 &  0.490 \\
resizedOutTileH(b0!=0) &  0.274 &  0.738 &  0.791 &  0.776 &  1.343 &  0.257 &  0.260 &  0.256 &  0.027 &  0.024 &  0.024 \\
sizeIntTileH &  0.384 &  0.246 &  0.264 &  0.259 &  0.269 &  0.514 &  0.519 &  0.256 &  0.270 &  0.474 &  0.490 \\
resizedInTileH(b0==0) &  0.274 &  1.230 &  0.527 &  0.518 &  0.215 &  0.026 &  0.026 &  0.077 &  0.054 &  0.024 &  0.024 \\
resizedInTileH(b0!=0) &  0.055 &  0.098 &  0.105 &  0.104 &  0.161 &  0.026 &  0.026 &  0.026 &  0.001 &  0.000 &  0.000 \\
\bottomrule
\end{tabular}%
}
\label{tab:hyper_f_score_normalized}
\end{table}

\subsection{Extended experimental results for all layers of ResNet18}

\begin{figure}[!h]
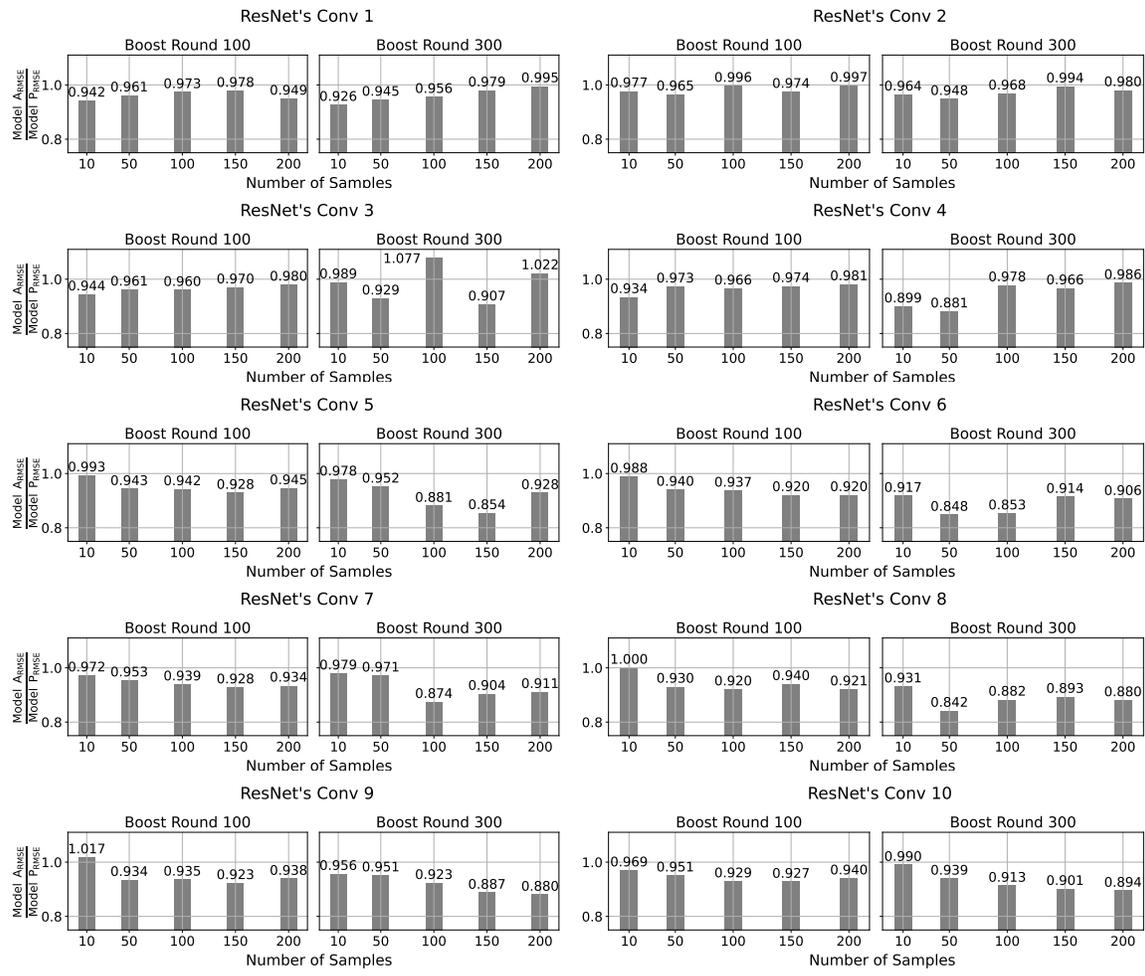

\centering    
\foreach \i in {1,3,5,7,9} {
    \begin{minipage}{1\textwidth}
        \hspace*{-1cm}
        \begin{minipage}{0.47\textwidth}
            \centering
            \includegraphics[width=1.2\textwidth]{Paper/Resources/pdf/\i.pdf}
            \vspace{-0.75cm}
        \end{minipage}%
        \hspace{0.08\textwidth} 
        \begin{minipage}{0.47\textwidth}
            \centering
            \includegraphics[width=1.2\textwidth]{Paper/Resources/pdf/\the\numexpr\i+1\relax.pdf}
            \vspace{-0.75cm}
        \end{minipage}
    \end{minipage}
}
\vspace{0.5cm}
\caption{Ratio of RMSE values of model A compared to model P per layers}
\label{fig:comparison_between_system_and_tvm}
\end{figure}



Fig.~\ref{fig:comparison_between_system_and_tvm} presents a comparison of the root mean squared error (RMSE) ratios between Model A and Model P across different layers, highlighting the impact of varying the number of boosting rounds and configuration samples. To compute the RMSE values displayed on the Y-axis, performance metrics were collected for all possible parameter configurations executable on the VTA.
The dataset was split into training and test sets, with the training set generated by {\systemName} based on the specified number of configuration samples. 
To reduce experimental error, each experiment was repeated 10 times, and the average results were calculated.

The results for Models P and A were plotted as functions of the number of configuration samples and the number of boosting rounds in XGBoost. In these graphs, the Y-axis represents the ratio of RMSE values between $\text{RMSE}_{\text{Model P}}$ and $\text{RMSE}_{\text{Model A}}$, while the X-axis indicates the number of configuration samples.


The results indicate that, for most layers, Model A achieves higher accuracy on the test set after training compared to Model P. Notably, increasing the number of boosting rounds from 100 to 300 improves test set accuracy. Specifically, the average test set accuracy increases from 0.916 with 100 boosting rounds to 0.932 with 300 boosting rounds.


\begin{figure}[h]
\centering   
\includegraphics[width=1\textwidth]{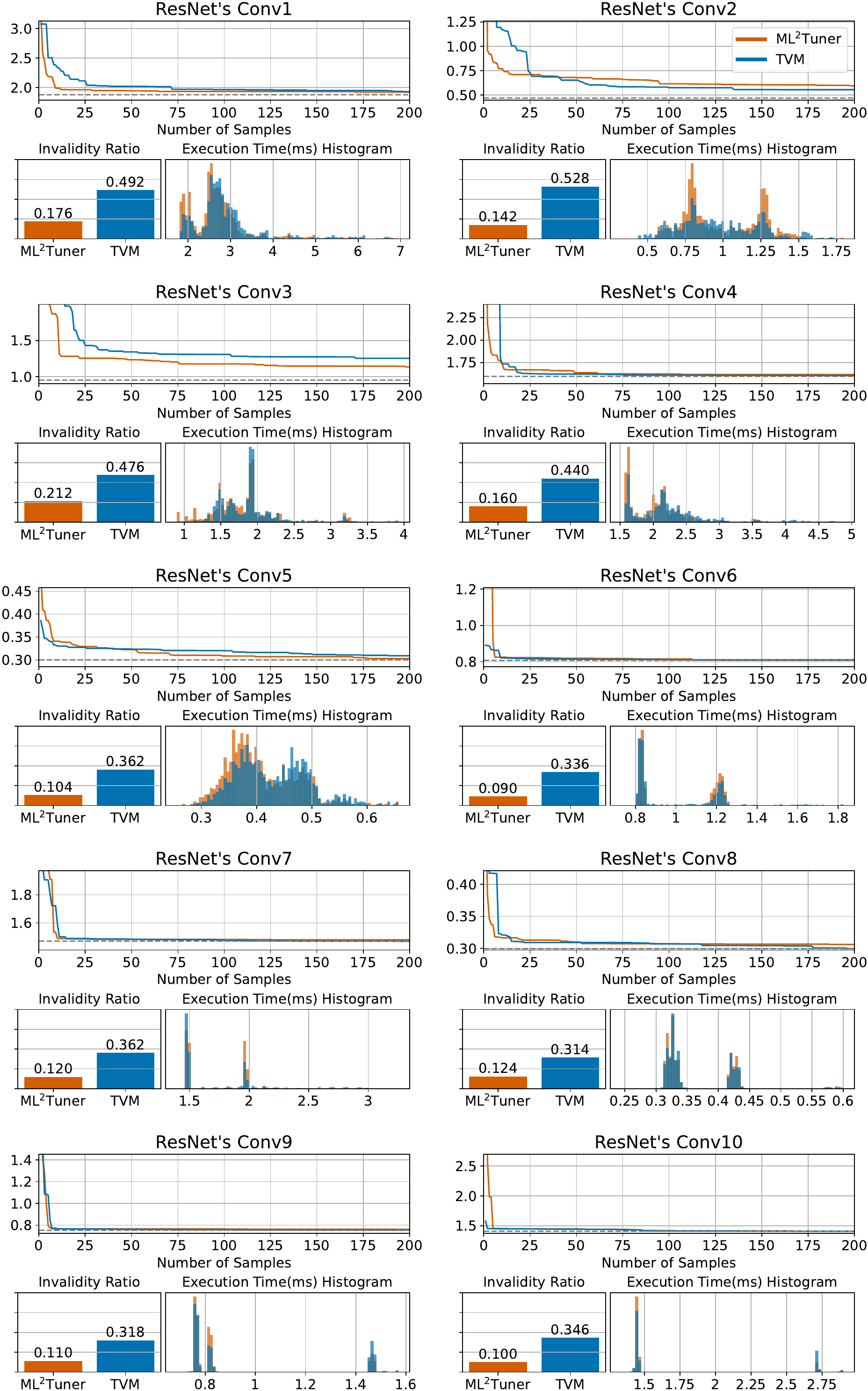}
\caption{Result of ResNet18 per layer on \systemName}
\label{fig:result_of_resnet18_all}
\end{figure}


Fig.~\ref{fig:result_of_resnet18_all} illustrates the results
of ResNet's convolution layers, emphasizing the effects of differences between validity prediction and advanced performance prediction. Layers 1 through 10 show that fewer invalid configurations are explored compared to the TVM method. Notably, {\systemName} achieves higher performance with fewer profiling attempts, particularly in Conv1 through Conv3, where the invalidity ratio is higher, thus outperforming TVM in terms of efficiency.

\end{document}